%% file: main.tex
\documentclass[sigconf]{acmart}
\copyrightyear{2021}
\acmYear{2021}
\acmConference[WSDM WebTour'21]{ACM WSDM Workshop on Web Tourism (WSDM Webtour'21}{March 12, 2021}{Jerusalem, Israel}
\acmBooktitle{ACM WSDM Workshop on Web Tourism (WSDM Webtour'21), March 12, 2021, Jerusalem, Israel}
\acmPrice{}
\acmDOI{}
\acmISBN{}
\usepackage{algorithm,algpseudocode}
\usepackage{amsmath}
\usepackage{amsfonts}
\usepackage{color}
\usepackage{listings}
\usepackage{caption}

\begin{document}

\title{Attention-based neural re-ranking approach for next city in trip recommendations}


\author{Aleksandr Petrov}
\affiliation{%
  \city{Edinburgh}
  \country{United Kingdom}}
\email{firexel@gmail.com}

\author{Yuriy Makarov}
\affiliation{%
  \city{Saint-Petersburg}
  \country{Russia}}
\email{lvoursl@gmail.com}

\renewcommand{\shortauthors}{Petrov and Makarov}

\begin{abstract}
    This paper describes an approach to solving the next destination city recommendation problem for a travel reservation system. We propose a two stages approach: a heuristic approach for candidates selection and an attention neural network model for candidates re-ranking. Our method was inspired by listwise learning-to-rank methods and recent developments in natural language processing and the transformer architecture in particular.
    We used this approach to solve the Booking.com recommendations challenge \cite{booking2021challenge}. Our team achieved 5\textsuperscript{th} place on the challenge using this method, with 0.555 accuracy@4 value on the closed part of the dataset. 
\end{abstract}





\maketitle

\section{Introduction}
\input{introduction}

\section{Attention-Based Reranking model}
\input{model}

%page3 

%page4

\section{Evaluation and Results} % Sasha

\input{evaluation_results}

\section{Conclusion}
As this challenge has shown, combining the listwise ranking methods and neural models is a viable idea that allowed us to achieve a high score on the contest. The same optimization approach can be used with other architectures and solve many other ranking problems in the information retrieval area, including recommender systems and search results ranking.

\bibliographystyle{ACM-Reference-Format}
\bibliography{references}

\end{document}

%% file: introduction.tex
\subsection{Problem description}
Next city in a trip recommendations is an important applied problem. When a user plans their trip, the ability to correctly predict next destination can directly benefit the user, saving them time on planning. The benefit of the user is important for the booking service provider - if the user gets precise recommendation, they are more likely to book it using the same platform and therefore increase the company revenue. 
In late 2020 Booking.com, the largest hotel reservations service provider, released a dataset and launched a public competition \cite{booking2021challenge} with the goal of achieving the best Accuracy@4 metric. 
We built an attention based model for this competition and achieved 5\textsuperscript{th} place with final score of  0.555 Accuracy@4 on the closed test set. 

\subsection{Dataset and target metric}
The dataset, released by booking.com, consists of two parts: \textit{train} and \textit{test}. Both parts include anonymized hotel checkins with following features: userId, checkinDate, checkoutDate, cityId, deviceClass, affiliateId, bookerCountry, hotelCountry, utripId. CityId and hotelCountry of the last checkin in the trip were masked in the \textit{test} part of the dataset and were used for scoring by the competition organizers. 
\\
The goal of the competition was to generate recommendations list of 4 cities. The organizers used Accuracy@4 as the main contest metric, which essentially is a percentage of times when our recommendations contained correct cityId. 

%% file: model.tex
Accuracy@4 is a ranking metric; therefore, we solved the problem as a \textit{ranking} problem. Accuracy@4 is not a very stable metric, as it does not take into account the order of the elements in the recommendations list and takes into account only 4 scores for each recommendation. Instead, we chose another popular ranking metric - NDCG@40\cite{ndcg} as our main optimization target. This metric is more stable as it considers the order of the elements in the list and takes into account 40 scores rather than 4. 

We chose the 2-stages approach, which is popular for solving ranking problems. In the first stage, we selected 500 candidates from all cities using a mixture of simple models. In the second stage, we re-ranked the candidates using a self-attention neural network. We used LambdaRANK\cite{burges2010ranknet} approach for the optimization, as it optimizes the NDCG metric directly. 

Our main model was inspired by the transformer architecture \cite{vaswani2017attention}. The architecture is designed for language processing and achieved state-of-the-art results in many text analysis tasks. The idea of utilizing a language model comes from the fact that a sequence of cities in the trip has a very similar structure as a sequence of words in the text. This idea was already successfully applied to recommender systems; see \cite{bert4rec} for example.  

On the high level, the model works as follows: 
\begin{enumerate}
\item Select candidates for ranking. 
\item Generate candidates matrix $C$. Each row of this matrix contains a vector representation of the corresponding candidate. The vector consists of learnable embeddings as well as engineered features. 
\item Generate vector representations of all known cities in the trip $T$ using transformer-like architecture
\item Generate vector representation of the last city in the trip $F$. Even though the target city itself is unknown, the dataset contains such features as checkinDate, checkoutDate, deviceClass, affiliateId, bookerCountry, which we can use for the prediction. 

\item Generate scores for the candidates. To do this we use MultiHeadAttention mechanism \cite{vaswani2017attention} between Candidate Cities and Trip and then calculate the final score using dot product with the encoded features vector:
\begin{equation}
    Scores = MultiHeadAttention(C, T) \cdot F
\end{equation}
\end{enumerate}

%page 2

\subsection{Labels generation}
The model training process requires labeled data. To get the labels for each trip, we performed the following procedure: 
\begin{enumerate}
    \item Sort actions in the trip by the time of check-in. 
    \item Choose a random fraction of the last cities in the trip as target cities. We re-split the data on each training epoch. This technique allowed us to generate multiple training samples out of a single trip. 
    \item Assign label score $2^{-i}$ for $i^{th}$ city in the target fraction. The idea is that each city from the target fraction is relevant as a trip continuation, but we want our model to rank higher the cities that the user visited in the near term. The intuition behind giving exponentially diminishing weights for future trips is based on the fact that the cities that the user will visit in the short-term future are more likely to be related to the cities the user visited recently. 
\end{enumerate}
\subsection{Candidates selection}

Our model uses the LamdaRANK ranking optimization approach\cite{burges2010ranknet}, which involves heavy computations and only can generate scores for a limited amount of candidate cities at once. Given that, we needed to generate good candidates using some simpler approaches. To train the model reasonably fast, we limited the number of candidates to 500 (out of 39870). We used a mixture of simpler models to fill the set of candidates. 

\paragraph{Basic models for candidates selection}
\label{section:candidate_selection_models}

This paragraph describes our basic models and heuristics that we used to produce a list of the candidate cities. During each training epoch, we randomly chose 10000 trips to train our main neural model to prevent overfitting, and we used the rest of the training set to re-train baseline models and generate features. Here is the list of the models that we used to generate candidates:
\begin{enumerate}
    \item \textit{All cities from the trip}. We have found that quite frequently, the user already visited the last city from the trip previously in the same trip, so all the cities from the trip are good candidates. We added all the cities from the trip to the set of candidates.

    \item \textit{TransitionChain} --  this model utilizes a sequential nature of the data.\\
    Let's assume that the total number of cities in the dataset is equal to $M$, and each trip has $K$ cities, where $K$-th city is a target city. We can create transition matrix $T \in \mathcal{R}^{M\times M}$, 
     which we can fill in a way, described in Algorithm \ref{TransitionsChain}.\\
     The prediction process is straightforward: we take all cities in the trip (except target) and sum up all lines in the matrix corresponding to these cities, then we took cities with the highest scores and used them as predictions. 
     
    We generated recommendations for the trip using TransitionsChain and added the resulting recommendations into the candidates set until it reached size 150. 
    
    \makeatletter
    \def\BState{\State\hskip-\ALG@thistlm}
    \makeatother
    
    \begin{algorithm}
    \caption{Transition Matrix ($T$) Filling}\label{TransitionsChain}
    \begin{algorithmic}[1]
    \Procedure{fillTransitionMatrix}{}
        \State $\textit{T} \gets \text{zero matrix with size M} \times{M}$
        \For{\emph{current\_trip in trips}}:
            \State ${last\_city} \gets {current\_trip[-1]}$
            \State ${prev\_cities} \gets {current\_trip[:-1]}$
            \For{\emph{city in prev\_cities}}:
                \State ${T[city][last\_city]} \mathrel{{+}{=}} {1}$.
            \EndFor
        \EndFor
    \EndProcedure
    \end{algorithmic}
    \end{algorithm}
    
    \item \textit{BookerTripCountryTop} -- This model generates the most popular cities in trips from the users from a particular country who had the same country of the last city in the trip. We added recommendations from this model to the candidates set until we got 350 candidate cities. 
    
    \item \textit{LastCityCountryTop} -- This model calculates the most popular next city based on the country of the last city in the trip and generates them as recommendations. We added recommendations from this model to the candidates set until we got 500 candidate cities.

    In some cases, we were not able to fill candidates set using the models above. For example, when a user visited very unpopular cities, we don't have enough statistics in the training set. In this case, we added candidates to the candidates set from two additional models: 

    \item \textit{BookerCountryTop} -- This model calculates most popular cities among users from a specific country. We added candidates from this model until we have 500 candidates in the set.

    \item \textit{GlobalTop} -- This model generates the most popular cities in the training set as recommendations. We added candidates from this model if all previous models could not fill the set of candidates. 
\end{enumerate}

Our experiments have shown that this heuristic includes the right candidate into the set of candidates with 90\% probability.

\subsection{Generate candidate's features}
\label{section:candidates_features}
For each candidate, we generated a vector representation that included the following features: 
%Describe features from booking_candidates_recommender.py -- Ura
\begin{itemize}
    \item City's popularity globally and especially for booker's country 
    \item City's popularity according to current month and year
    \item City's score from TransitionsChain recommender.
    \item Binary flag: is the candidate same city as the first city in the trip
    \item Binary flags: for each of the last 5 cites in the trip - is the candidate the same as this city from the trip 
    \item Candidate's cosine similarity with last 5 cities in a trip based on their co-occurrence in train data.
\end{itemize}

\subsection{Trip Encoding}
\label{section:trip_encoding}
Our model used transformer\cite{vaswani2017attention}-like architecture to encode cities from the trip history. 
We represented each city in the trip as a vector, that included following parts:
\begin{enumerate}
    \item City embedding - 32 dimensions, learnable.
    \item Booker country embedding - 32 dimensions, learnable.
    \item Hotel country embedding - 32 dimensions, learnable (share weights with the booker country embedding)
    \item Affiliate id embedding - 5 dimensions, learnable.
    \item City-in-trip features. Manually engineered features, that include: 
    \begin{itemize}

        \item Number of nights between check-in and check-out dates
        \item Is this trip over the weekend or not?
        \item Is the current city in the same country as previous ones?
        \item City's check-in and check-out day of week and day of the year
        \item Is booker's country equal to hotel's country?
        \item Check-in year (3 features, one-hot encoded)
        \item Checkin month (12 features, one-hot encoded)
    \end{itemize}
\end{enumerate}

Overall, each city in the trip was represented using 115 parameters. We stacked the cities' representations in the trip and padded the matrix to the shape (50, 115), where 50 is the maximum possible number of cities in the trip. 

To get a better semantic representation of the cities in the trip, we encoded each city using a single dense layer with 115 dimensions.  

Similarly to the original transformer architecture, we combined city representation with positional embeddings. In the original paper\cite{vaswani2017attention} the authors use fixed embeddings, based on $sin$ function. We replaced this with two learnable embeddings, representing the city's position from the beginning and the end of the trip. These embeddings are then concatenated, passed through a dense layer, and multiplied with city representation vectors. 

To model interactions of the cities in the trip, we added three transformer-like blocks. The original transformer block consists of multi-head attention, feed-forward layer, residual connection, and layer normalization. Our transformer-like block has a similar architecture with only a minor difference: we used Multiplication for residual connection instead of Sum. Our experiments have shown that the network with Multiplication instead of Sum required fewer epochs to train. Figure \ref{img:transformer_block} shows architecture of the Transformer-like block used in our architecture. 
\begin{figure}[h!]
    \begin{center}
        \item \mbox{}
    \includegraphics[width=0.27\textwidth]{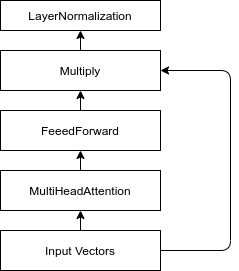}    
    \end{center}
    \caption{Transformer-like block}
    \label{img:transformer_block}
\end{figure}

We then used the Transformer-like part of the model to model interactions between candidate cities and cities from the trip. In addition to the features matrix described in section \ref{section:candidates_features}, we added city embedding and country embedding to each candidate. These embeddings share weights with the city embeddings and country embeddings from the trip-encoding part of the model. The candidates were additionally encoded using a feed-forward layer. We also added one transformer-like block to model the relationship between candidates. The idea is that the candidates are not independent, and the model should score candidates knowing about other candidates. A similar  idea of modeling interactions between candidates for recommendations is described in \cite{pei2019personalized}

We used one multi-head attention layer to model interaction between encoded user trips and encoded candidate cities. We hypothesize that this output layer contained semantically rich representations of the (trip, candidate) pairs. 

% attention -Sasha
\begin{figure}[h!]
\begin{center}
    \item \mbox{}
\includegraphics[width=0.37\textwidth]{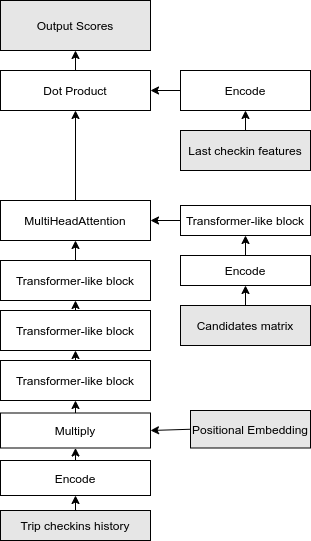}    
\end{center}
\caption{High-level model architecture}
\label{img:architecture}
\end{figure}

To generate the model's final scores, we used dot product operation with encoded features of the target city. The dataset contains all parameters of the target city, except cityId and countryId. We encoded these parameters using the same manually engineered features as described in section \ref{section:candidates_features} and encoded them using a dense layer. 

\subsection{Model training}
Our model's goal was to rank candidates according to the label score, which constitutes a ranking task. The problem of direct optimization of the ranking metrics, such as Accuracy@4 or NDCG, is difficult because these metrics only depend on the order of the items and therefore are not smooth and can not be directly optimized gradient-descent technologies. To overcome this problem, we used the LambdaRANK approach described in \cite{burges2010ranknet}. The method's idea is that instead of calculating the loss function and calculating its gradients, one can directly calculate gradients. The authors of \cite{burges2010ranknet} propose generating such gradients for optimizing the NDCG metric. We implemented this method and used it as our main optimization approach. 

We optimised our network using Adam\cite{kingma2014adam} optimiser. We trained our model using 50-epochs early stop criteria on val\_accuracy\_at\_4 metric. 
\begin{figure}[h!]
\begin{center}
    \item \mbox{}
\includegraphics[width=0.48\textwidth]{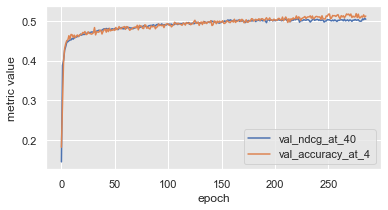}    
\end{center}
\caption{Validation performance during training process}
\label{img:transformer_block}
\end{figure}

Figure \ref{img:transformer_block} illustrates model performance on the validation set during the training process. As one can see from the plot, the NDCG@40 (metric optimized by the LambdaRANK approach) and Accuracy@4 (contest target metric) demonstrate high correlation, suggesting that optimizing one metric leads to optimization of the other. 

 We used Keras and TensorFlow\cite{tensorflow} libraries to train our neural networks. Training of the model took 7 hours on GeForce GTX3090 GPU. 

%% file: evaluation_results.tex
In this section, we describe the validation scheme, metrics and analyze our results. 

\subsection{Validation scheme}
Our validation scheme was based on trips. We used both \textit{train} and \textit{test} datasets provided by the competition organizers to train our model.  We randomly allocated 4000 trips to the hold-out set, 4000 trips to the validation set, and the rest to the training set. We tried to predict the last city in the trip based on all previous cities' information and available information about the last city in the validation and hold-out sets. 

\subsection{Metrics}
We used the following metrics to evaluate our models:
\begin{itemize}
\item{\verb|Accuracy@4|}: competition organizers proposed this metric. The idea of Accuracy@4 is quite simple: it is equal to 1 if the target city in our top-4 prediction and 0 otherwise. 
\item{\verb|NDCG@40|}: according to our experiments, NDCG@40 is a more stable metric than previous ones. We also found that NDCG@40 and Accuracy@4 have a quite strong correlation (Figure 3.). One can find the details about the NDCG metric in \cite{ndcg}
\item{\verb|Leaderboard|}: this metric is essentially the Accuracy@4 metric, calculated by the organizers on the closed part of the dataset. We could only calculate this metric for two models by the competition's conditions - the one that we sent for the intermediate leaderboard and the final submission. 
\end{itemize}

\subsection{Model performance against baselines}

We used the following baselines to evaluate quality of our model:
\begin{itemize}
    \item \textit{GlobalTop, LastCityCountryTop, TransitionChain} - baselines used as part of our candidates selection process (see section \ref{section:candidate_selection_models})
    \item \textit{TruncatedSVD} - a popular method of solving recommendations problem \cite{koren2009matrix}. 
    
    \item \textit{SelfAttention} - end-to-end version of our model.  The model's architecture consists of a transformer-like part of our model, followed by three dense layers, the last of which generates recommendation scores for all cities in the dataset. The main difference compared with the reranking attention model is that this model doesn't use manually engineered candidate features. This makes the model simpler, and therefore the model can produce scores for all cities in the dataset rather than a small number of candidates. 
    
    \item \textit{LambdaMART} -  Gradient boosting trees implementation of the LambdaRANK approach. We used the same candidate generation procedure and same manually engineered features described in section \ref{section:candidates_features}. We used LambdaMART implementation from LightGBM lilbrary\footnote{https://lightgbm.readthedocs.io/en/latest/}. 
\end{itemize}

\begin{table}
  \caption{Models comparison}
  \label{tab:results}
  \begin{tabular}{cccl}
    \toprule
    Model&Accuracy@4&NDCG@40&Leaderboard\\
    \midrule
    GlobalTop & 0.058 & 0.091 & -\\
    TruncatedSVD & 0.261 & 0.261 & - \\
    LastCityCountryTop & 0.372 & 0.358 & -\\
    TransitionChain & 0.440 & 0.429 & -\\
    SelfAttention & 0.509 & 0.491 & 0.514\\
    LambdaMART & 0.514 & 0.485 & -\\
    \textbf{Reranking Attention}  & \textbf{0.542} & \textbf{0.513} & \textbf{0.555}\\
  \bottomrule
\end{tabular}
\end{table}

Table \ref{tab:results} contains results of the comparison. Our final model demonstrates statistically significant improvement over baselines on the Accuracy@4 metric (p-value < 0.01 in two-tailed Z-test). 

%% file: main.bbl

\begin{thebibliography}{9}

%%% the Web address.

\ifx \showCODEN    \undefined \def \showCODEN     #1{\unskip}     \fi
\ifx \showDOI      \undefined \def \showDOI       #1{#1}\fi
\ifx \showISBNx    \undefined \def \showISBNx     #1{\unskip}     \fi
\ifx \showISBNxiii \undefined \def \showISBNxiii  #1{\unskip}     \fi
\ifx \showISSN     \undefined \def \showISSN      #1{\unskip}     \fi
\ifx \showLCCN     \undefined \def \showLCCN      #1{\unskip}     \fi
\ifx \shownote     \undefined \def \shownote      #1{#1}          \fi
\ifx \showarticletitle \undefined \def \showarticletitle #1{#1}   \fi
\ifx \showURL      \undefined \def \showURL       {\relax}        \fi
% invisible to TeX
\providecommand\bibfield[2]{#2}
\providecommand\bibinfo[2]{#2}
\providecommand\natexlab[1]{#1}
\providecommand\showeprint[2][]{arXiv:#2}

\bibitem[\protect\citeauthoryear{Abadi, Agarwal, Barham, Brevdo, Chen, Citro,
  Corrado, Davis, Dean, Devin, Ghemawat, Goodfellow, Harp, Irving, Isard, Jia,
  Jozefowicz, Kaiser, Kudlur, Levenberg, Man\'{e}, Monga, Moore, Murray, Olah,
  Schuster, Shlens, Steiner, Sutskever, Talwar, Tucker, Vanhoucke, Vasudevan,
  Vi\'{e}gas, Vinyals, Warden, Wattenberg, Wicke, Yu, and Zheng}{Abadi
  et~al\mbox{.}}{2015}]%
        {tensorflow}
\bibfield{author}{\bibinfo{person}{Mart\'{\i}n Abadi}, \bibinfo{person}{Ashish
  Agarwal}, \bibinfo{person}{Paul Barham}, \bibinfo{person}{Eugene Brevdo},
  \bibinfo{person}{Zhifeng Chen}, \bibinfo{person}{Craig Citro},
  \bibinfo{person}{Greg~S. Corrado}, \bibinfo{person}{Andy Davis},
  \bibinfo{person}{Jeffrey Dean}, \bibinfo{person}{Matthieu Devin},
  \bibinfo{person}{Sanjay Ghemawat}, \bibinfo{person}{Ian Goodfellow},
  \bibinfo{person}{Andrew Harp}, \bibinfo{person}{Geoffrey Irving},
  \bibinfo{person}{Michael Isard}, \bibinfo{person}{Yangqing Jia},
  \bibinfo{person}{Rafal Jozefowicz}, \bibinfo{person}{Lukasz Kaiser},
  \bibinfo{person}{Manjunath Kudlur}, \bibinfo{person}{Josh Levenberg},
  \bibinfo{person}{Dandelion Man\'{e}}, \bibinfo{person}{Rajat Monga},
  \bibinfo{person}{Sherry Moore}, \bibinfo{person}{Derek Murray},
  \bibinfo{person}{Chris Olah}, \bibinfo{person}{Mike Schuster},
  \bibinfo{person}{Jonathon Shlens}, \bibinfo{person}{Benoit Steiner},
  \bibinfo{person}{Ilya Sutskever}, \bibinfo{person}{Kunal Talwar},
  \bibinfo{person}{Paul Tucker}, \bibinfo{person}{Vincent Vanhoucke},
  \bibinfo{person}{Vijay Vasudevan}, \bibinfo{person}{Fernanda Vi\'{e}gas},
  \bibinfo{person}{Oriol Vinyals}, \bibinfo{person}{Pete Warden},
  \bibinfo{person}{Martin Wattenberg}, \bibinfo{person}{Martin Wicke},
  \bibinfo{person}{Yuan Yu}, {and} \bibinfo{person}{Xiaoqiang Zheng}.}
  \bibinfo{year}{2015}\natexlab{}.
\newblock \showarticletitle{{TensorFlow}: Large-Scale Machine Learning on
  Heterogeneous Systems}.
\newblock  (\bibinfo{year}{2015}).
\newblock
\urldef\tempurl%
\url{https://www.tensorflow.org/}
\showURL{%
\tempurl}
\newblock
\shownote{Software available from tensorflow.org.}


\bibitem[\protect\citeauthoryear{Burges}{Burges}{2010}]%
        {burges2010ranknet}
\bibfield{author}{\bibinfo{person}{Christopher~JC Burges}.}
  \bibinfo{year}{2010}\natexlab{}.
\newblock \showarticletitle{From ranknet to lambdarank to lambdamart: An
  overview}.
\newblock \bibinfo{journal}{\emph{Learning}} \bibinfo{volume}{11},
  \bibinfo{number}{23-581} (\bibinfo{year}{2010}), \bibinfo{pages}{81}.
\newblock


\bibitem[\protect\citeauthoryear{Goldenberg, Kofman, Levin, Mizrachi, Kafry,
  and Nadav}{Goldenberg et~al\mbox{.}}{2021}]%
        {booking2021challenge}
\bibfield{author}{\bibinfo{person}{Dmitri Goldenberg}, \bibinfo{person}{Kostia
  Kofman}, \bibinfo{person}{Pavel Levin}, \bibinfo{person}{Sarai Mizrachi},
  \bibinfo{person}{Maayan Kafry}, {and} \bibinfo{person}{Guy Nadav}.}
  \bibinfo{year}{2021}\natexlab{}.
\newblock \showarticletitle{Booking.com WSDM WebTour 2021 Challenge}.
  \bibinfo{howpublished}{\url{https://www.bookingchallenge.com/}}. In
  \bibinfo{booktitle}{\emph{ACM WSDM Workshop on Web Tourism (WSDM
  WebTour’21)}}.
\newblock


\bibitem[\protect\citeauthoryear{J{\"a}rvelin and
  Kek{\"a}l{\"a}inen}{J{\"a}rvelin and Kek{\"a}l{\"a}inen}{2002}]%
        {ndcg}
\bibfield{author}{\bibinfo{person}{Kalervo J{\"a}rvelin} {and}
  \bibinfo{person}{Jaana Kek{\"a}l{\"a}inen}.} \bibinfo{year}{2002}\natexlab{}.
\newblock \showarticletitle{Cumulated gain-based evaluation of IR techniques}.
\newblock \bibinfo{journal}{\emph{ACM Transactions on Information Systems
  (TOIS)}} \bibinfo{volume}{20}, \bibinfo{number}{4} (\bibinfo{year}{2002}),
  \bibinfo{pages}{422--446}.
\newblock


\bibitem[\protect\citeauthoryear{Kingma and Ba}{Kingma and Ba}{2014}]%
        {kingma2014adam}
\bibfield{author}{\bibinfo{person}{Diederik~P Kingma} {and}
  \bibinfo{person}{Jimmy Ba}.} \bibinfo{year}{2014}\natexlab{}.
\newblock \showarticletitle{Adam: A method for stochastic optimization}.
\newblock \bibinfo{journal}{\emph{arXiv preprint arXiv:1412.6980}}
  (\bibinfo{year}{2014}).
\newblock


\bibitem[\protect\citeauthoryear{Koren, Bell, and Volinsky}{Koren
  et~al\mbox{.}}{2009}]%
        {koren2009matrix}
\bibfield{author}{\bibinfo{person}{Yehuda Koren}, \bibinfo{person}{Robert
  Bell}, {and} \bibinfo{person}{Chris Volinsky}.}
  \bibinfo{year}{2009}\natexlab{}.
\newblock \showarticletitle{Matrix factorization techniques for recommender
  systems}.
\newblock \bibinfo{journal}{\emph{Computer}} \bibinfo{volume}{42},
  \bibinfo{number}{8} (\bibinfo{year}{2009}), \bibinfo{pages}{30--37}.
\newblock


\bibitem[\protect\citeauthoryear{Pei, Zhang, Zhang, Sun, Lin, Sun, Wu, Jiang,
  Ge, Ou, et~al\mbox{.}}{Pei et~al\mbox{.}}{2019}]%
        {pei2019personalized}
\bibfield{author}{\bibinfo{person}{Changhua Pei}, \bibinfo{person}{Yi Zhang},
  \bibinfo{person}{Yongfeng Zhang}, \bibinfo{person}{Fei Sun},
  \bibinfo{person}{Xiao Lin}, \bibinfo{person}{Hanxiao Sun},
  \bibinfo{person}{Jian Wu}, \bibinfo{person}{Peng Jiang},
  \bibinfo{person}{Junfeng Ge}, \bibinfo{person}{Wenwu Ou}, {et~al\mbox{.}}}
  \bibinfo{year}{2019}\natexlab{}.
\newblock \showarticletitle{Personalized re-ranking for recommendation}. In
  \bibinfo{booktitle}{\emph{Proceedings of the 13th ACM Conference on
  Recommender Systems}}. \bibinfo{pages}{3--11}.
\newblock


\bibitem[\protect\citeauthoryear{Sun, Liu, Wu, Pei, Lin, Ou, and Jiang}{Sun
  et~al\mbox{.}}{2019}]%
        {bert4rec}
\bibfield{author}{\bibinfo{person}{Fei Sun}, \bibinfo{person}{Jun Liu},
  \bibinfo{person}{Jian Wu}, \bibinfo{person}{Changhua Pei},
  \bibinfo{person}{Xiao Lin}, \bibinfo{person}{Wenwu Ou}, {and}
  \bibinfo{person}{Peng Jiang}.} \bibinfo{year}{2019}\natexlab{}.
\newblock \showarticletitle{BERT4Rec: Sequential Recommendation with
  Bidirectional Encoder Representations from Transformer}. In
  \bibinfo{booktitle}{\emph{Proceedings of the 28th ACM International
  Conference on Information and Knowledge Management}} (Beijing, China)
  \emph{(\bibinfo{series}{CIKM '19})}. \bibinfo{publisher}{ACM},
  \bibinfo{address}{New York, NY, USA}, \bibinfo{pages}{1441--1450}.
\newblock
\showISBNx{978-1-4503-6976-3}
\urldef\tempurl%
\url{https://doi.org/10.1145/3357384.3357895}
\showDOI{\tempurl}


\bibitem[\protect\citeauthoryear{Vaswani, Shazeer, Parmar, Uszkoreit, Jones,
  Gomez, Kaiser, and Polosukhin}{Vaswani et~al\mbox{.}}{2017}]%
        {vaswani2017attention}
\bibfield{author}{\bibinfo{person}{Ashish Vaswani}, \bibinfo{person}{Noam
  Shazeer}, \bibinfo{person}{Niki Parmar}, \bibinfo{person}{Jakob Uszkoreit},
  \bibinfo{person}{Llion Jones}, \bibinfo{person}{Aidan~N Gomez},
  \bibinfo{person}{Lukasz Kaiser}, {and} \bibinfo{person}{Illia Polosukhin}.}
  \bibinfo{year}{2017}\natexlab{}.
\newblock \showarticletitle{Attention is all you need}.
\newblock \bibinfo{journal}{\emph{arXiv preprint arXiv:1706.03762}}
  (\bibinfo{year}{2017}).
\newblock


\end{thebibliography}
